\documentclass[times,twocolumn,final,authoryear]{elsarticle}
\usepackage[twocolumn,textwidth=18cm,columnsep=.5cm]{geometry}

\usepackage{framed,multirow}

\usepackage{amssymb}
\usepackage{latexsym}
\usepackage{amsmath}

\usepackage[linesnumbered,ruled,vlined]{algorithm2e}

\SetCommentSty{mycommfont}

\makeatletter
\newcommand{\removelatexerror}{\let\@latex@error\@gobble}
\makeatother

\usepackage{url}
\usepackage{xcolor}
\definecolor{newcolor}{rgb}{.8,.349,.1}

\usepackage{xparse}

\usepackage{array}
\usepackage{hhline}
\usepackage{makecell}
\newcolumntype{M}[1]{>{\centering}m{#1}}

\journal{Pattern Recognition Letters}

\title{LV-ROVER: Lexicon Verified Recognizer Output Voting Error Reduction}

\author{Bruno STUNER, Cl\'ement CHATELAIN, Thierry PAQUET} 

\begin{document}

\begin{abstract}
Offline handwritten text line recognition is a hard task that requires both an efficient optical character recognizer and language model. 
Handwriting recognition state of the art methods are based on Long Short Term Memory (LSTM) recurrent neural networks (RNN) coupled with the use of linguistic knowledge.
Most of the proposed approaches in the literature focus on improving one of the two components and use constraint, dedicated to a database lexicon.
However, state of the art performance is achieved by combining multiple optical models, and possibly multiple language models with the Recognizer Output Voting Error Reduction (ROVER) framework.
Though handwritten line recognition with ROVER has been implemented by combining only few recognizers because training multiple complete recognizers is hard.
In this paper we propose a Lexicon Verified ROVER: LV-ROVER, that has a reduce complexity compare to the original one and that can combine hundreds of recognizers without language models.
We achieve state of the art for handwritten line text on the RIMES dataset.
\end{abstract}

\maketitle

\section{Introduction}

Handwriting recognition is a difficult task due to the variability of shapes (the different writing styles) and the complexity of the language.
State of the art systems use dynamical models that provide a global recognition result at the line level. 
They combine optical character models with language models to achieve low Character Error Rates (CER). 
For a long time, systems combining Hidden Markov Models with Neural Networks were achieving the best performance\citep{plotz2009markov,bharath2012hmm}. 
The neural network was dedicated to the optical model of characters possibly with a HMM, while  the language model was implemented using statistical n-grams.
Recent developments have highlighted the superiority of Long Short Term Memories (LSTM) Recurrent Neural Networks (RNN) \citep{hochreiter1997long} trained with the Connectionist Temporal Classification (CTC) \citep{CTC} in place of the traditional shallow hybrid HMM/NN optical model.
Although recurrent neural networks implicitly learn character sequences (thus modeling linguistic information), they are still unable to achieve the full recognition task alone. 
Thus the need to combine them with language models, such as N-grams in a lexicon and language driven decoding scheme (N-best Viterbi decoding plus rescoring the hypotheses using high order language models in a second pass).
Moreover, the results of the recent READ 2016 competition \citep{sanchez2016icfhr2016} have shown that combining multiple recognition systems in a ROVER (Recognizer Output Voting Error Reduction) scheme \citep{fiscus1997post}, increases the performance. 
To summarize we can highlight that the actual state of the art performance is achieved by cascading a two stages combination and voting scheme that combines multiple optical models, and possibly multiple language models. 
However, training multiple complete recognizers is hard, due to the difficulty to train complementary optical character models and complementary linguistic models. 
This fact explains why handwritten line recognition with ROVER has been implemented by combining few recognizers, a few dozen as reported in \citep{bluche2014comparison,sanchez2016icfhr2016,bertolami2005ensemble}.

These observations let us think that the ROVER framework may be underexploited, leaving room to explore other voting schemes that could exploit in a better way the raw performance of RNN optical models that have significantly contributed to the performance increase observed these last years. 
In this paper we are interested in developing a ROVER scheme that can support the combination of hundreds, even thousands of raw recognizers (without language models). In this purpose, we propose a modified version of the ROVER algorithm that allows fast alignment of many recognizer outputs, by avoiding the computational burden of aligning multiple recognizers outputs using dynamic programming.
Our proposition also extends the traditional voting scheme by the integration of a Lexicon Verification Voting scheme. 
A second contribution of the paper lies in the introduction of a training trick that allows getting many complementary raw BLSTM recognizers (cohort of BLSTM) in a very effective way.  
This new ROVER scheme achieves state of the art performance on the RIMES dataset without the need for any language model.

This article is organized as follows: in section 2, state of the art in handwriting recognition and ROVER combination is reviewed, and the principle of LSTM RNN is recalled. Then in section 3, we present our new ROVER algorithm. Section 4 is dedicated to the presentation of the training trick that allows the generation of a cohort of BLSTM. Thereafter we present the experimental results in section 5 before concluding.

\section{Related works}

In this section, the state of the art of handwriting recognition is reviewed. The Recognizer Output Voting Error Reduction (ROVER) system is also presented and analyzed.

\subsection{Handwriting recognition}

Handwriting recognition is the process of transcribing images of handwritten texts into strings of characters.
The process is usually carried on following two steps, the optical character recognition (OCR) and the linguistic processing.

Nowadays, almost\footnote{apart some preliminary works from a2ia \citep{bluche2016scan} that directly process paragraphs using attention-based deep neural network models} every handwriting recognition systems take as input pre-segmented text lines.
Text lines were then historically segmented into words and then into characters to train optical character recognizer. 
Today state-of-the art OCR methods directly recognize lines in their whole, using  Long Short Term Memory (LSTM) recurrent neural networks (RNN) trained with the Connectionist Temporal Classification \citep{graves2009offline}.
Despite the significant progress achieved with RNN, they cannot achieve state of the art performance alone. 
Similar to the approaches introduced some decades ago with Hidden Markov Models \citep{rabiner1989tutorial,plotz2009markov,bharath2012hmm}, there is a need to combine the RNN optical models with lexicons and language models in a lexicon and language model driven decoding scheme to achieve the state of the art recognition performance \cite{hamdani2013open,poznanski2016cnn,voigtlaender2016handwriting}.

Although state-of-the art systems now provide interesting performance \cite{grosicki2009icdar}, the idea of combining several recognizers has emerged, in order to reach better performance.
However, combining the output of several recognizers is not trivial, mainly due to the dynamical nature of the models, that may lead to output signals of different sizes.
The most popular combination method is the ROVER method \citep{fiscus1997post} that aligns results of many systems (each operating a lexicon and language model driven decoding scheme), and votes for the most probable solution.
ROVER has become a popular solution introduced in many methods of the literature  \citep{bluche2014comparison,sanchez2016icfhr2016,zamora2014neural}
or with some modification \citep{wang2002combination,bluche2015limsi,xu2011minimum}.
We now present the ROVER combination scheme in the next paragraph.

\subsection{ROVER combination}

For a good description of the Recognizer Output Voting Error Reduction (ROVER) scheme, it is worth going back to the original paper by Fiscus (1997), and refer to figure \ref{rover} below. 
As stated by the author:

 \begin{quote}{\cite{fiscus1997post}}
The  ROVER system  is implemented  in  two modules.  First, the system  outputs  from two  or more ASR systems  are  combined into  a single  word  transition network. [...] Once  the  network  is  generated,  the  second  module  evaluates  each  branching  point  using  a  voting  scheme,  which  selects  the  best  scoring word (with the highest number  of  votes) for  the  new transcription.
\end{quote}

\begin{figure}[h]
\includegraphics[width=\columnwidth]{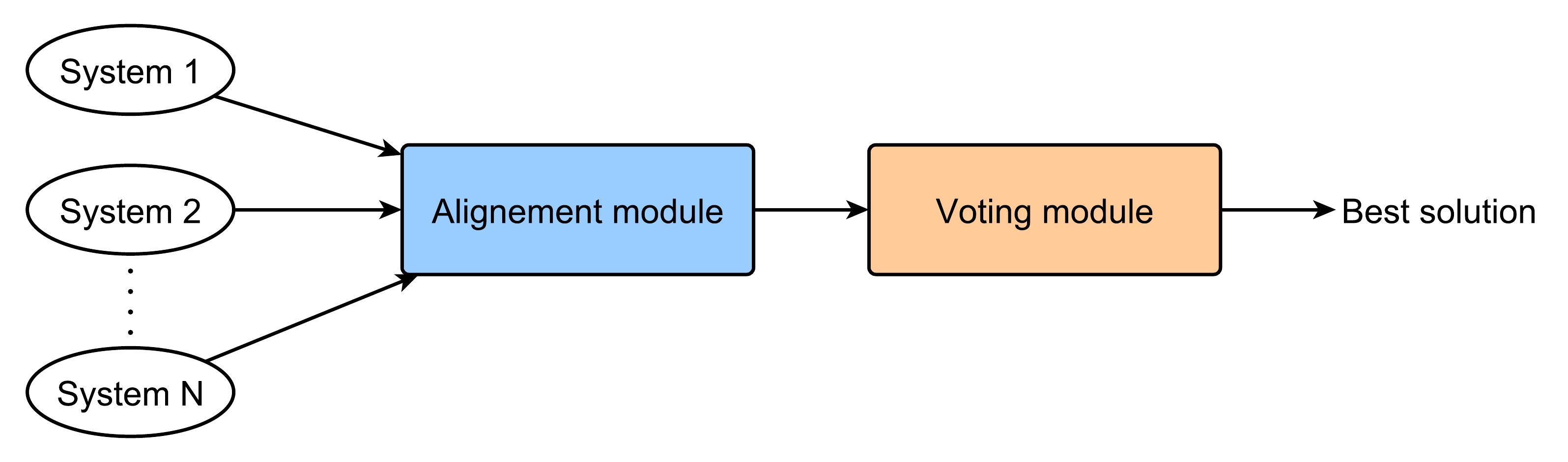}
\caption{The ROVER process as presented in \citep{fiscus1997post}}
\label{rover}
\end{figure}

As depicted in figure \ref{rover}, the ROVER scheme is divided 
in two modules: 1- the alignment module 2- the voting module. 
Several voting scheme are presented in the original article \citep{fiscus1997post}, whereas only one  alignment procedure is exposed.
The alignment consists in producing a Word Transition Networks (WTN) by using the classical Dynamic Programming (DP) algorithm \citep{bellman1956dynamic}.
Finding the optimal alignment for n sequences has been shown to be NP-complete \citep{wang1994complexity}.
Thus, as mentioned by J. G. Fiscus, dynamic programming can not be implemented with such an hyper-dimensional search.
This is due to its exponential cost and the curse of dimensionality \citep{bellman2015adaptive}.
This is the reason why ROVER has been implemented using an approximation of the best alignment. 
The proposed alignment use the dynamic programming to align  all sequences against a selected reference sequence, thereby reducing the complexity of the search into two dimension. 
This algorithm produce a Word Transition Network.
Even if the alignment is proceed with an approximate algorithm, the ROVER complexity remains high.
The ROVER alignment complexity is $O(N\times{}l\times{}L\times{}L')$ where $N$ is the number or recognizers, $L$ the number of words in the reference sequence,  $L'$ the number of words in an other sequence and $l$ the length of the output (i.e. its number of characters).

The ROVER voting module is running on the word transition network produced by the alignment module.
It consists in a vote regarding the frequency of occurrence and if considered a score of the recognizer.
The search is performed in the WTNs by considering transition arcs as independent.
The complexity of this search is $O(N\times{}L)$.

In addition to the complexity of the ROVER combination scheme, one must also consider the complexity of each recognizer, which generally operates in a two pass decoding scheme to allow decoding with high order language models.
This is probably the reason why the ROVER combination has always been carried out with a small number of recognizers, no more than a few dozen to our knowledge, and always by including both the optical model and the language model  \citep{bluche2014comparison,sanchez2016icfhr2016,bertolami2005ensemble}.

All in all, we can see that successful implementations of the ROVER combination scheme require intensive computation in many respects, whereas the performance gain is moderate in general, although always present.
Also, the literature review shows that very little is said about how to get  complementary systems that could be combined favorably using ROVER.
This is why we believe that ROVER like combination scheme has been  underexploited until now due to the many limitations we have highlighted above, which can be summarized by complexity of the combination scheme and low complementarity of the systems. 
In the next paragraph we propose a new low complexity ROVER combination algorithm: LV-ROVER, that supports the combination of hundreds of systems.

\section{Lexicon Verified ROVER: LV-ROVER}

\begin{figure*}[h!]
\centering
\includegraphics[width=0.85\paperwidth]{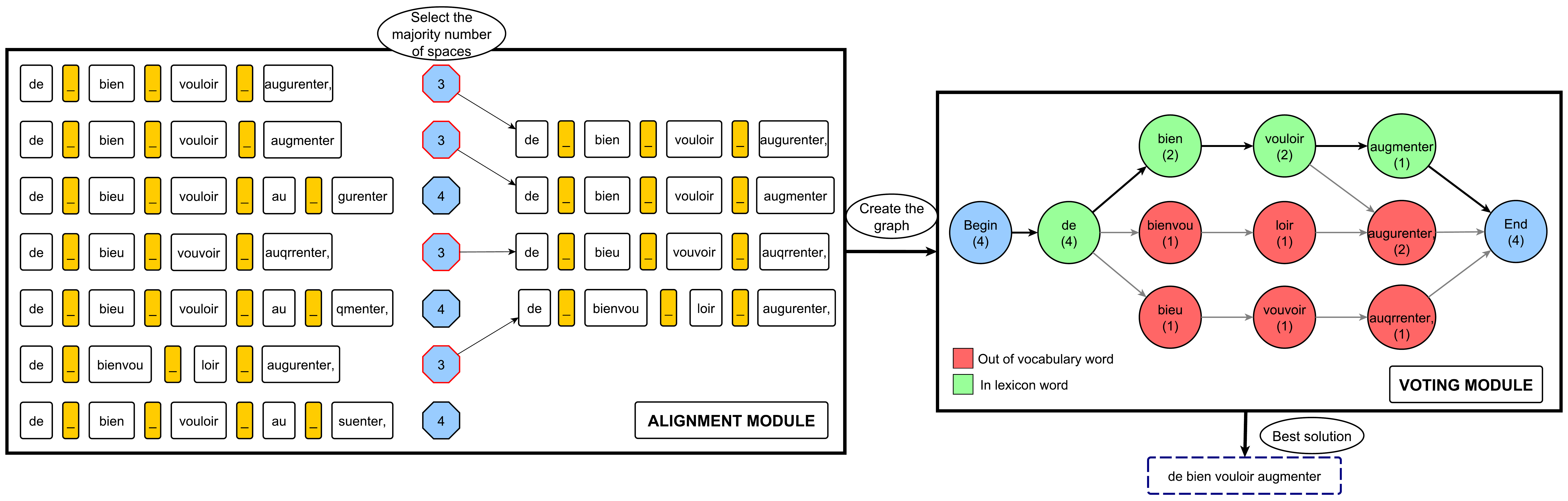}
\caption{Lexicon Verified Recognizer Outputs Voting Error Reduction.}
\label{combiLigne}
\end{figure*}

In this work, we propose a new method LV-ROVER that performs similarly or better to the standard ROVER but with a reduced complexity, that can combine hundreds of RAW recognizers like LSTM RNN. 
As the complexity of both the alignment and the voting module of ROVER are relatively high, there is a need for changing their principle.
LV-ROVER is based on majority voting and lexicon verification.
First the alignment is carried out by counting the number of spaces in each recognition results allowing to generate a words lattice of every recognition results having the majority number of spaces.
Secondly, the voting module uses a lexicon verification rule in order to get high confidence word hypotheses when searching for the best sequence of words in the graph.
LV-ROVER combination is pictured on figure \ref{combiLigne} with an example showing how each module works.

\subsection{Alignment module}

The alignment module in the ROVER framework is used to align multiple recognition results at line level. 
However, as seen previously the ROVER alignment module is based on a dynamic programming algorithm (DP) which can only provide an approximation of the alignment between systems outputs, and this is performed with a high complexity.
In our case, combining hundreds of recognizers with no language model like LSTM RNN, induces two main weak difficulties for the DP alignment:
First there are character errors in the word hypotheses, making the matching process between line hypotheses prone to these errors.
Second, only dynamic programming can be used as hyper dimensional DP is intractable because of it's exponential complexity and the curse of dimensionality \citep{bellman2015adaptive}.

Considering these two limitations only an approximate solution can be considered. 
In this respect, we propose to split the alignment module in two sequential steps. 
In the first step, the number of words of the final sequence is estimated.
This had a direct effect on the complexity of the alignment module. 
This is simply performed by proceeding to a majority vote for the number of spaces in the sequence over the pool of output sequences.
Word count estimation is simply performed by majority voting over the pool of output sequences.
In a second step, words are aligned by considering only the outputs matching the word count estimate.
These selected outputs are combined into a word lattice, with a fixed number of words.
The words lattice is then passed to the voting module.

The proposed alignment, while allowing to be more flexible regarding character errors than dynamic programing, has a lower complexity as it is only $O(N\times{}(l))$ where $N$ is the number of recognizer outputs and $l$ is the number of characters in the recognizer outputs sequence.
The algorithm \ref{algo1} describes the voting procedure for alignment. 

\begin{algorithm}
\SetKwInOut{Input}{Input}
\SetKwInOut{Output}{Output}
\Input{N = nbNetwork, inputStrings = [N]}
\Output{wordsLattice[N][]}
spaceCounter = [0,...,0] \;
\tcp{Iteration over the number of networks}
\For{$i = 0; i < N$}
{
spaceNumber = 0 \;
    currentString = ''\;
    \tcp{Counting the number of spaces in inputString}
\ForEach{$Character$ \textbf{in} $inputStrings[i]$}
{
\uIf{Character == '$\_$'}{
    wordsLattice[i][spaceNumber] = currentString\;
    currentString = ''\;
	spaceNumber $\leftarrow$ spaceNumber + 1\;
}
\uElse{currentString += Character;}
}
spaceCounter[spaceNumber] += 1\;
}
nbWords = argMax(spaceCounter)\;
\For{$i = 0; i < N$}
{\tcp{Only keeps sequence of nbWords}
\uIf{length(wordsLattice[i]) != nbWords}{
remove(wordsLattice[i])\;}
}
\Return wordsLattice, nbWords
\caption{LV-ROVER alignment voting procedure.}
\label{algo1}
\end{algorithm}

\subsection{Voting module}
The voting module is based on word frequencies as in \citep{fiscus1997post}.
However, we introduce a lexicon verification rule while exploring the words lattice.

Lexicon verification is the process that validates or rejects a word hypothesis produced by a lexicon free recognizer.
The verification simply validates a word hypothesis if it belongs to the lexicon.
On the opposite, if the word doesn't belong to the lexicon then it is discarded.
Lexicon verification has proved to be a reliable and efficient rule in a previous work \citep{Cohort16} for isolated words recognition.
In the ROVER combination scheme as lines are segmented into words, verification appears to be an efficient rule also.

The word lattice is then converted into a graph where connections are only made between words that were consecutive in at least one input string.
The graph is explored from the beginning to the end and vice versa.
The solution containing the highest number of lexicon verified words form both ways is selected.
For sake of clarity, we avoid the burden of the transformation to a particular optimized structured graph, we present the algorithm directly on the words lattice in algorithm \ref{algo2}.

\begin{figure*}[!t]
 \removelatexerror
\begin{algorithm*}[H]
\SetKwInOut{Input}{Input}
\SetKwInOut{Output}{Output}
\Input{lexicon, wordsLattice[N'][nbWords] \tcp{$N'$ is the number of remaining networks}}  
\Output{resultString}
nextSolutions = wordsLattice[:][0]\;
\For{$j \leftarrow 0$ to $nbWords$}{
\tcc{Create two new empty dictionaries (which are hashtables): inLexiconWords and oovWords. They have for key the word and for values the frequency of the word and a dictionary containing every possible next solution.}
inLexiconWords $\leftarrow$ new dictionary\{\}[[]]\; 
\tcc{Access the dictionary values with inLexiconWords\{word\} and values[0] being the frequency and values[1] the dictionary containing next solutions. The same applies for oovWords.}
oovWords $\leftarrow$ new dictionary\{\}[[]]\;
  \For{$i \leftarrow 0$ to $N'$ }{
	word = wordsLattice[i][j]; 
    nextWord = wordsLattice[i][j+1]\;
    \tcc{Check if the word is in the next solutions of the previous word as to not make segmentation errors.}
    \uIf{word \textbf{in} nextSolutions}{
      \uIf{word \textbf{in} lexicon}{
          \uIf{word \textbf{in} inLexiconWords}{
              inLexiconWords\{word\}[0] += 1; \tcp{Increment frequency}    
              inLexiconWords\{word\}[1].add(nextWord); \tcp{Add next word to next solutions}
          }
          \uElse{
          	  inLexiconWords.add(word)\;
              inLexiconWords\{word\}[0] $\leftarrow$ 1; \tcp{Initialize frequency to 1}
              inLexiconWords\{word\}[1].add(nextWord); \tcp{Add next word to next solutions}
          }
      }
      \uElse{
          \uIf{word \textbf{in} oovWords}{
              oovWords\{word\}[0] += 1\;
              oovWords\{word\}[1].add(nextWord)\;
          }
          \uElse{
          	  oovWords.add(word)\;
              oovWords\{word\}[0] $\leftarrow$ 1\;
              oovWords\{word\}[1].add(nextWord)\;
          }      
      }
    }
  }
  \uIf{inLexiconWords is not empty}{
  	resultString += maxFrequency(inLexiconWords)\;
    nextSolutions $\leftarrow$ inLexiconWords\{maxFrequency(inLexiconWords)\}[1];
  }
  \uElse{  	
  	resultString += maxFrequency(oovWords)\;
    nextSolutions $\leftarrow$ oovWords\{maxFrequency(oovWords)\}[1];
  }
}
\Return resultString
\caption{Lexicon verification voting procedure.}
\label{algo2}
\end{algorithm*}
\end{figure*}

The complexity of this search is $O(2\times{}N'\times{}W)$ where $N'$ is the number of remaining classifiers outputs and $W$ the estimated number of words in the alignment module.
This set of voting rules is fast and very efficient through the words lattice and allows finding a quasi-optimal solution.
It also allows to handle Out Of Vocabulary words, which is a strong advantage in comparison to lexicon driven decoding methods.

In the following section, we present the experimentations carried out to 
assess this effectiveness of the proposed LV-ROVER algorithm. 
One of the most important aspect of the successful experimentation results lies in the generation of hundreds of complementary LSTM RNN.

\section{Datasets and cohort generation}

\subsection{Datasets and lexicon\label{lexicons}}

In our experiments we use the Rimes dataset \citep{grosicki2011icdar}.

The Rimes dataset is composed of 11328 training lines and 778 evaluation lines.
We use 3 different lexicons for this tasks:
\begin{itemize}
\item The standard Rimes lexicon containing all the words from the training data (8578 words);
\item The French dictionary Gutenberg (FDG) (336 531 words);
\item A gigantic lexicon (GIG) containing all the French words from the French Wikipedia and French Wiktionnaire pages (3 276 994 words).
\end{itemize}

\subsection{Cohort generation \label{cohort}}

Generating a large ensemble of complementary networks is difficult.
The most common ways for getting complementarity are very limited.
For example, different architectures of the networks or different input features could be used.
However, it is costly in terms of design and training time.
Another known way to generate complementary networks is to use different random initializations as proposed in \citep{menasri2012a2ia}, however it is still costly in training time.

\begin{figure}[h]
\includegraphics[width=\columnwidth]{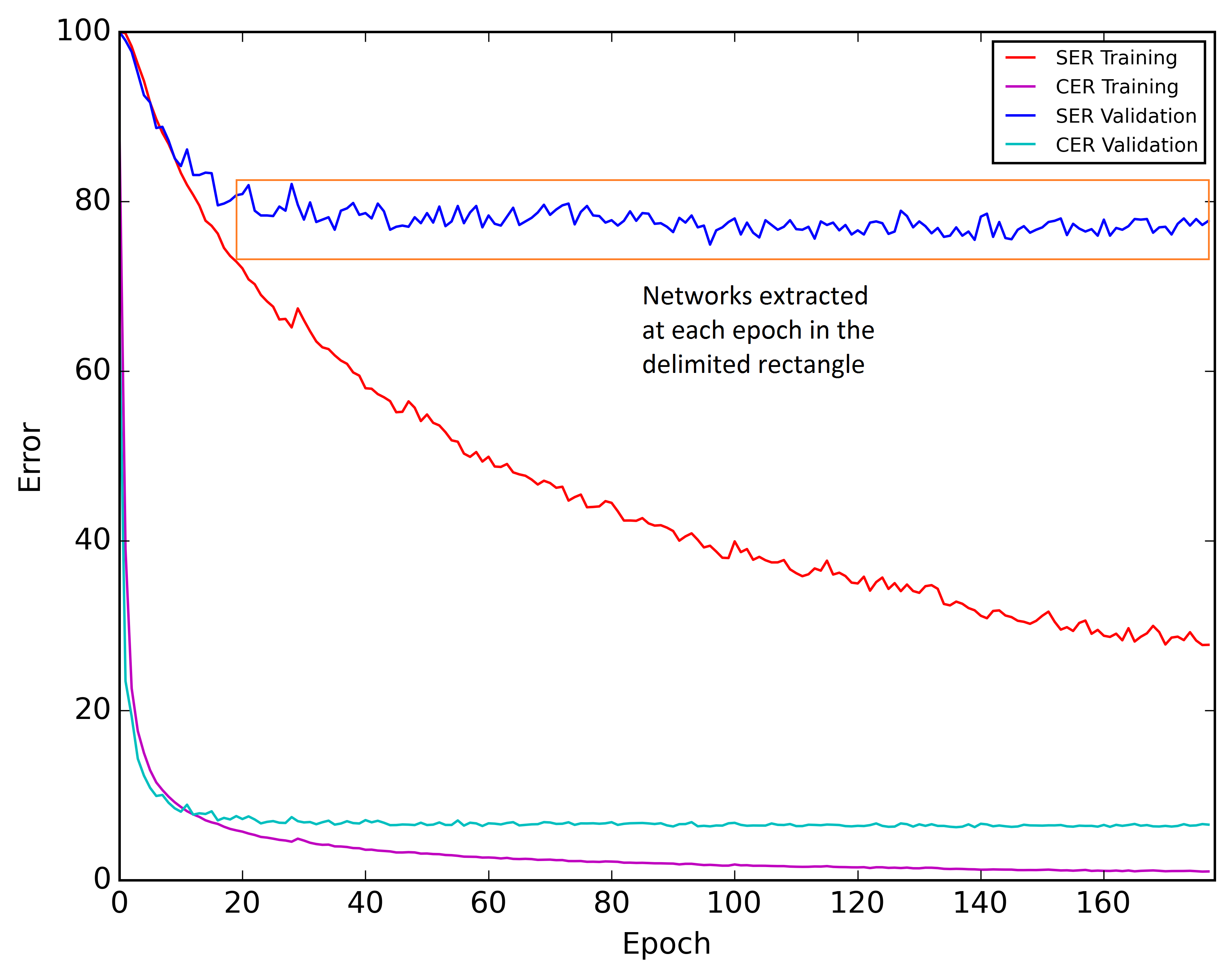}
\caption{Training curves of a single LSTM RNN from which the cohort is extracted presenting the Character Error Rate (CER) and the Sequence Error Rate (SER).}
\label{cohortCurv}
\end{figure}

To overcome these limitations, we proposed in \citep{Cohort16} to extract hundred of networks in a single training procedure.
The rationale behind this approach relies on the theoretical work of \cite{choromanska2015loss} which concludes that the loss function of a fully-connected neural network converges towards multiple local minima with equivalent magnitude.
Thus, selecting every network in these local minima may provide a large number of complementary networks with different weights that may perform equally as shown on figure \ref{cohortCurv}.
The network ensemble extracted by this strategy is called a cohort.
Our earlier work demonstrated the efficiency of the approach, however it requires the training parameters to be specifically adapted so as to avoid being trapped in a local minimum.
To obtain complementary networks while training one single network, the following parameters must be selected:
\begin{itemize}
\item Shuffle the data between each epoch ;
\item Use momentum to escape local minimum, as it is common in literature 0.9 ;
\item Set a learning rate neither too big (no convergence) neither too small (trapped in local minimum) like $10^{-4}$.
\end{itemize}

In order to get more complementary networks, we also decided to train four different networks on the Rimes dataset. They are as follows:
\begin{itemize}
\item One MDLSTM with 3 layers of 2, 10, 50 LSTM;
\item Two BLSTM with 3 layers of 60,70,120 LSTM with different initializations;
\item One BLSTM with 3 layers of 80,100,120 LSTM.
\end{itemize}
In total these four trainings yields 454 complementary LSTM RNN.

All the trainings were performed with the RNNLIB \citep{graves2013rnnlib}.
We now present the experimental results in the next section and show that while performing faster, LV-ROVER performs better to combine hundreds of LSTM RNN without any language model in comparison to the original ROVER.

\section{Results}

The results obtained by the LV-ROVER are presented in two parts. First the study of the LV-ROVER performance regarding the lexicon size and its contribution against the classical ROVER framework is analyzed, then a comparison of our results to other states of the art methods is made.
Two metrics in the results section are used: the Character Error Rate (CER) and the Word Error Rate (WER).

\subsection{LV-ROVER performance}

It is well known that the lexicon size of a handwriting recognition method greatly influences the recognition results \citep{koerich2003large}. 
A large lexicon makes the recognition harder (confusion between words are more likely), at the expend of the dataset coverage.
We present in table \ref{lex} the results of the LV-ROVER approach on the line Rimes dataset, recognized using the various lexicons described in section \ref{lexicons}, whose sizes vary from 8.6K to 3.3M words.
As our approach can work in a lexicon-free fashion, we also provide the results without any lexicon. 
One can observe that increasing the lexicon size has only a small impact on performance: the LV-ROVER CER performance slightly increases when adding the French Dictionary to the Rimes lexicon, and it increases again when adding the gigantic lexicon to the Rimes one. 
It shows that our approach supports extremely large lexicons, while being very efficient.
However when using only the gigantic lexicon, CER stays stable and WER is increased by  only 0.23\%. 
It is due to the coverage loss due to the absence of the dedicated lexicon. 
These very interesting results using extremely large lexicons open new perspective for recognizing handwriting without any dedicated lexicons.

\begin{table}[h]
  \centering
  \begin{tabular}{|M{2cm}M{1cm}|M{1.3cm}|M{1.1cm}|M{1.1cm}|}
  \hline
  \textbf{Lexicon} & Lex.\\size & \textbf{Coverage\\(\%)} & \textbf{CER (\%)} & \textbf{WER\\(\%)} \tabularnewline
\Xhline{0.5mm}
  No lexicon & 0 & 0 & 4.06 & 15.76  \tabularnewline    
  Rimes & 8.6K & 95.23 & 2.53 & 7.84  \tabularnewline
  Rimes + FDG & 342K & 97.59 & 2.74 & 8.18 \tabularnewline    
  Rimes + GIG & 3.3M & 98.65 & 3.16 & 10.42  \tabularnewline
  GIG 	& 3.3M & 97.28 & 3.17 & 10.65  \tabularnewline   
  \hline
  \end{tabular}
  \caption{Comparison of the results of LV-ROVER on the Rimes line test dataset, using various lexicon (See section \ref{lexicons} for lexicon description).}
  \label{lex}
\end{table}

We now compare the results of LV-ROVER to the classical ROVER algorithm.
Even if our LV-ROVER approach has a lower complexity than the ROVER framework, both approaches are able to combine hundreds of LSTM RNN.
Thus in table \ref{ourVsRover}, the results of the combination of both LV-ROVER and the classic ROVER are presented, using the 454 networks described in section \ref{cohort}, on the Rimes line test dataset.
One can observe that the LV-ROVER with the Rimes lexicon performs better than the classical ROVER. 
This is also true for other lexicons as presented before in table \ref{lex}.
The classic ROVER is under performing due to the fact that in either case no language model is used, and as a consequence the alignment is harder for the ROVER system as recognitions results are not smoothed by the language model.
This suggests that the ROVER method is best suited for combining results of complete recognitions systems including both optical character recognition and language model, whereas LV-ROVER manage combining raw recognizers thanks to its alignment and lexicon verification.

\begin{table}[h]
  \centering
  \begin{tabular}{|M{3.5cm}|M{1.4cm}|M{1.4cm}|}
  \hline
  \textbf{Method} &  \textbf{CER(\%)} & \textbf{WER(\%)} \tabularnewline
\Xhline{0.5mm}
  ROVER & 4.73 & 16.74 \tabularnewline     
  LV-ROVER (this work)&  \textbf{2.53} & \textbf{7.84}  \tabularnewline
  \hline
  \end{tabular}
  \caption{Comparison of ROVER and LV-ROVER with the Rimes lexicon results using 454 recognizers on the Rimes test dataset.}
  \label{ourVsRover}
\end{table}

In conclusion the LV-ROVER has a very low sensitivity to the lexicon size and therefore can be used with very large lexicons.
Any lexicons can be used, provided it fits the language, but no necessarily the dataset.
Moreover it outperforms the classic ROVER method for combining hundreds of raw LSTM RNN.
We now compare the results of our LV-ROVER approach to other state of the art methods.

\subsection{Literature results comparison}

In table \ref{ourVsEA} we compare the LV-ROVER to other state of the art methods on the Rimes dataset. 
Our system is the same presented in the previous subsection, i.e. the LV-ROVER combination of the 454 networks.
In \citep{pham2014dropout} the authors use a MDLSTM RNN with dropout and a 4-gram language model, in \citep{voigtlaender2016handwriting} they use a similar approach with a large MDLSTM CNN and a 4-gram word based language model.
We outperform both methods regarding both CER and WER, the improvement is especially high for the WER.
Moreover if we were to compare our results with the French dictionary, the LV-ROVER is still performing better than other state of the art methods.

\begin{table}[h]
  \centering
  \begin{tabular}{|M{4cm}|M{1.2cm}|M{1.2cm}|}
  \hline
  \textbf{Method} &  \textbf{CER} & \textbf{WER} \tabularnewline
\Xhline{0.5mm}
  \cite{voigtlaender2016handwriting} &  2.8 & 9.6  \tabularnewline    
  \cite{pham2014dropout} &  3.3 & 12.3 \tabularnewline      
  LV-ROVER (this work)&  \textbf{2.5} & \textbf{7.8}  \tabularnewline

  \hline
  \end{tabular}
  \caption{Performance comparison of our method against states of the art methods on the Rimes dataset.}
  \label{ourVsEA}
\end{table}

\section{Conclusion}

This work presents the Lexicon Verified ROVER (LV-ROVER) a new ROVER-like combination, that allows to combines hundreds of recognizers outputs.
The combination is build around two modules: the alignment module which aligns outputs regarding the number of words, and the voting module which is based on lexicon verification.
Both modules prove their efficiency for segmenting and voting while being of a lesser complexity than ROVER.
State of the art results are achieved on the Rimes dataset.
Future works can extend the low sensitivity to the lexicon to allow recognition of images of different languages with a multilingual lexicon.

\bibliographystyle{model2-names}
\bibliography{refs}

\end{document}